\documentclass[conference]{IEEEtran}
\IEEEoverridecommandlockouts

\usepackage[numbers]{natbib}

\usepackage{amsmath}
\usepackage{amssymb}
\usepackage{amsfonts}
\usepackage{graphicx}
\usepackage{booktabs}
\usepackage{enumitem}
\usepackage{url}
\usepackage{microtype}
\usepackage{pifont}
 \usepackage{tikz}
\usetikzlibrary{arrows.meta,positioning,calc}
\newcommand{\sysname}{SciTraj}
\usepackage{multirow}

\begin{document}

\title{How Does Research Evolve? Tracing Cross-Domain Trajectories in NLP, ML, and CV through a Claim-Grounded Typed Citation Corpus}

\author{
\IEEEauthorblockN{
Abdul Muntakim\IEEEauthorrefmark{1},
Md Abdullah Al Hafiz Khan\IEEEauthorrefmark{1},
Sadid Hasan\IEEEauthorrefmark{2},
Yong Pei\IEEEauthorrefmark{1}
}
\IEEEauthorblockA{
\IEEEauthorrefmark{1}Kennesaw State University, Georgia, USA\\
\IEEEauthorrefmark{2}Microsoft, Cambridge, USA\\
amuntaki@students.kennesaw.edu,
\{mkhan74, ypei\}@kennesaw.edu,
sadidhasan@microsoft.com
}
}

\maketitle

\begin{abstract}

How does research evolve, and can we trace it at the level of individual claims? Scientific progress is not simply a uniform accumulation of facts: ideas extend prior methods, address known limitations, realize proposed future directions, and sometimes dispute earlier claims. Existing citation graphs usually collapse these roles into a single homogeneous edge type, limiting how we can analyze scientific progress.
We introduce \sysname{}, a typed citation corpus for tracing research evolution across natural language processing, machine learning, and computer vision. \sysname{} includes 32{,}559 papers published between 2015 and 2024 and 573{,}126 directed edges spanning six research-relation types. Unlike traditional citation graphs, each edge is paired with the claim sentence that motivates its label. Claim-driven relations are verified by natural language inference against their local in-paper context. The corpus further organizes these relations into multi-step typed trajectories that trace how ideas develop across papers and over time. We evaluate the corpus along three dimensions. First, a three-annotator pilot achieves Fleiss' $\kappa=0.74$ and 79.9\% majority-vote precision for relation labels, indicating substantial agreement and reliable labeling. Second, corpus-level analyses reveal clear disciplinary siloing in the directional flow of research relations. Topic analysis further identifies rapidly growing clusters dominated by vision and LLM-related research and declining clusters associated with several classical machine-learning topics. We further evaluate \sysname{} using a temporally split link-prediction benchmark and a year-shuffle falsifiability test that distinguishes genuine temporal signal from year-correlated content. Under this setting, \textsc{SciTraj-Pair} performs strongly, but its AUC drops by 0.288 when publication years are shuffled, showing that its predictions depend not only on content but also on the temporal order in which research develops. These results establish \sysname{} as an inspectable resource for studying how research relationships form and accumulate across papers, disciplines, and time, while providing a foundation for future research on scientific-trajectory modeling and forecasting.

\end{abstract}


\section{Introduction}
\label{sec:intro}

Scientific fields are often described as accumulating knowledge, but the process is rarely a uniform expansion of facts. Ideas extend earlier methods, address limitations that earlier authors themselves described, realize directions those authors proposed but did not pursue, and occasionally dispute what was previously published. This raises questions that are easy to ask and hard to answer precisely. Does a field advance mainly by extension or by correction? Do methodological ideas travel across neighbouring communities, or do they circulate largely within them? Given a research trajectory and the claims it has accumulated, can we say anything principled about where it is likely to go next?

A natural way to study how research develops is to examine the citation record itself. The scientific literature provides a direct trace of how a field evolves over time, but citation links are most useful when they capture \emph{why} one paper cites another. Large-scale resources such as S2ORC \citep{lo2020s2orc} and SPECTER2 \citep{singh2023scirepeval} capture citation structure and semantic similarity, while typed citation datasets \citep{jurgens2018measuring} assign citation-function labels; however, they do not directly identify the specific claim in the cited paper being addressed. To bridge this gap, we define a \emph{claim seed} as a sentence expressing a contribution, limitation, future direction, dispute, or causal dependence, and a \emph{research relation} as a directed, typed edge between two papers motivated by such a sentence. Anchoring relations to claim seeds makes each edge inspectable and supports claim-level analysis of multi-step research trajectories.

In this paper, we propose pairing each typed citation edge with the claim sentence that motivates it and verifying that claim against its local in-paper context. This creates a resource for studying research evolution directly and at scale. We develop this framework as \sysname{}, a typed, claim-seeded citation corpus spanning natural language processing, machine learning, and computer vision. For the four claim-driven relations, we retain only claims supported by their surrounding context. The two similarity-based relations are instead defined using abstract similarity and publication-year constraints. Our six relation types are designed to capture common forms of research continuation, response, and correction expressed in scientific writing.

We explore this idea concretely and empirically on 32{,}559 papers
published at major NLP, ML, and Vision venues between 2015 and 2024,
connected by 573{,}126 directed typed edges. Specifically, we:

\begin{enumerate}\itemsep0pt
\item Introduce a six-type directional research-relation schema that
captures distinct forms of research continuation, response, and correction (Section~\ref{sec:corpus_schema}).

\item Develop a claim-grounded extraction and verification pipeline that
links claim-driven relations to explicit claim sentences and verifies them
against their local in-paper context
(Section~\ref{sec:corpus_pipeline}).
  \item Introduce \sysname{}, a claim-grounded typed citation corpus with
 claim provenance and NLI verification for all
  claim-driven relations, comprising 32,559 papers and
  approximately 287.9M typed trajectories of length $\geq 3$ (Section~\ref{sec:corpus}).

  \item Ask (and answer) two questions about how research has evolved
  over the study window: how typed citation flow is distributed within
  and across the three communities, and which topics have emerged or
  declined, cross-validating both against the real Semantic Scholar
  citation subgraph (Section~\ref{sec:findings};
  Section~\ref{app:real_citations}).

  \item Propose a year-shuffle falsifiability protocol that separates
  genuine publication-time signal from year-correlated content, and
  establish a temporally split typed link-prediction benchmark
  (Section~\ref{sec:methods}; Section~\ref{sec:year_shuffle}).
\end{enumerate}

Our corpus-level analysis reveals two main patterns. First, research communities are strongly inward-looking: within-community citations occur more often than expected under a uniform-target baseline, while most cross-community flows fall below this baseline, with NLP$\leftrightarrow$Vision showing the strongest separation. The same directional pattern appears in the corresponding citation subgraph, suggesting that it is not an artefact of our edge-construction procedure. Second, between 2019 and 2024, emerging topics are concentrated in Vision and LLM-related research, while several classical ML topics show declining activity.

\sysname{} is intended to complement, rather than replace, LLM-based citation analysis. Although long-context models can examine individual paper pairs, they do not provide a versioned corpus with claim-level edge provenance, corpus-scale typed trajectories, or controlled temporal analyses such as year shuffling. We therefore position \sysname{} alongside prior work on typed citation classification, research-evolution analysis, and normalized cross-disciplinary research flow in Section~\ref{sec:related_work}.

\section{Related Work}
\label{sec:related_work}

\sysname{} brings together four closely related areas of research. These include typed citation classification, scientific claim extraction and verification, scientometric analysis of research evolution and cross-field flow, and link prediction on scholarly graphs.

Prior work such as ACL-ARC \citep{bird-etal-2008-acl} and SciCite \citep{cohan2019structural}, building on earlier citation-function research \citep{teufel-etal-2006-automatic}, classifies the rhetorical role of citations. Subsequent work has improved these models through graph-based representations \citep{berrebbi2022graphcite}, ensemble methods \citep{paolini2025citefusion}, and multi-sentence or multi-intent modeling \citep{lauscher-etal-2022-multicite}. These approaches explain how a citation functions in the citing text, but they do not directly identify the specific claim in the cited paper being acted upon, verify that relation against context, or compose relations across multiple papers. \sysname{} addresses these limitations through claim-level provenance, NLI filtering for claim-driven relations, and a directional relation schema that supports interpretable multi-step trajectories.

A complementary line of work focuses on extracting scientific claims and contributions. SciERC \citep{luan-etal-2018-multi} identifies entities and relations in scientific abstracts, while NLPContributionGraph \citep{dsouza-etal-2021-semeval} structures contribution-bearing statements as knowledge graphs. In parallel, FEVER and SciFact formalize claim verification through natural language inference over textual evidence \citep{thorne2018fever,wadden2020fact}. \sysname{} connects these directions by using extracted claims to seed typed inter-paper relations and applying NLI to determine whether each claim is supported by its surrounding context.

Research-evolution studies have examined changes in NLP through topic trends, paradigm shifts, and citation patterns \citep{prabhakaran-etal-2016-predicting,pramanick-etal-2023-diachronic}. Recent work also reports increasing disciplinary insularity and shifts toward more recent citations \citep{wahle-etal-2023-cite,wahle2025citation}. Such analyses reveal how research communities change and interact, but typically treat citation edges as homogeneous. \sysname{} complements this work by distinguishing the type of relationship represented by each edge and analyzing normalized typed flows across research communities.

Scholarly link prediction has progressed from structural heuristics \citep{adamic2003friends} to relational and temporal graph models \citep{schlichtkrull2018modeling,nguyen2023spatial}, often incorporating content representations such as SPECTER2 \citep{singh2023scirepeval}. However, evaluation remains sensitive to negative sampling and temporal split design \citep{poursafaei2022towards}, and most prior work predicts whether a citation will occur rather than the type of research relation it represents. \sysname{} provides a temporally split benchmark for typed relation prediction and introduces a year-shuffle test to examine whether predictive gains genuinely depend on temporal information rather than only on time-correlated content.


\section{The \sysname{} Corpus Curation}
\label{sec:corpus}

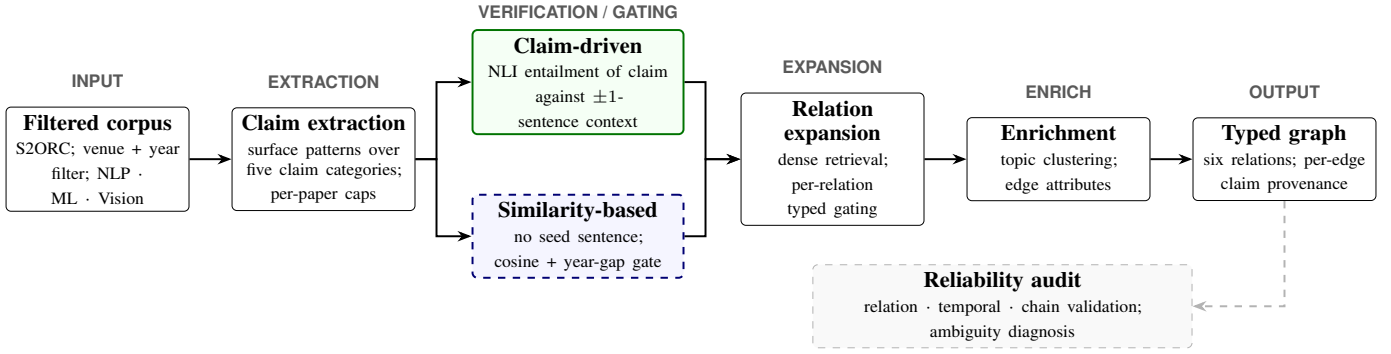
\begin{figure*}[t]
\centering
\resizebox{\textwidth}{!}{%
\begin{tikzpicture}[
  font=\small,
  box/.style={draw, rounded corners=2pt, align=center, fill=white,
              text width=24mm, inner sep=3pt},
  claim/.style={draw=green!45!black, thick, rounded corners=2pt,
                align=center, fill=green!4, text width=28mm, inner sep=3pt},
  simi/.style={draw=blue!45!black, thick, dashed, rounded corners=2pt,
               align=center, fill=blue!4, text width=28mm, inner sep=3pt},
  audit/.style={draw=gray!60, dashed, rounded corners=2pt, align=center,
                fill=gray!5, text width=52mm, inner sep=3pt},
  stage/.style={font=\scriptsize\bfseries\sffamily, text=black!65},
  ar/.style={-{Stealth[length=2mm]}, thick},
  arg/.style={-{Stealth[length=2mm]}, thick, gray!60, dashed}
]

\node[box] (corpus) {\textbf{Filtered corpus}\\[2pt]
  \scriptsize S2ORC; venue + year\\ filter; NLP $\cdot$ ML $\cdot$ Vision};

\node[box, right=6mm of corpus] (extract) {\textbf{Claim extraction}\\[2pt]
  \scriptsize surface patterns over\\ five claim categories;\\ per-paper caps};

\node[claim, right=8mm of extract, yshift=11mm] (nli) {\textbf{Claim-driven}\\[2pt]
  \scriptsize NLI entailment of claim\\ against $\pm1$-sentence context};

\node[simi, right=8mm of extract, yshift=-11mm] (sim) {\textbf{Similarity-based}\\[2pt]
  \scriptsize no seed sentence;\\ cosine + year-gap gate};

\node[box, right=8mm of nli.east, yshift=-11mm] (expand) {\textbf{Relation expansion}\\[2pt]
  \scriptsize dense retrieval;\\ per-relation typed gating};

\node[box, right=6mm of expand] (enrich) {\textbf{Enrichment}\\[2pt]
  \scriptsize topic clustering;\\ edge attributes};

\node[box, right=6mm of enrich] (out) {\textbf{Typed graph}\\[2pt]
  \scriptsize six relations; per-edge\\ claim provenance};

\node[audit, below=9mm of enrich, xshift=-8mm] (val)
  {\textbf{Reliability audit}\\[2pt]
   \scriptsize relation $\cdot$ temporal $\cdot$ chain validation;\\
   ambiguity diagnosis};

\node[stage, above=1.5mm of corpus]  {INPUT};
\node[stage, above=1.5mm of extract] {EXTRACTION};
\node[stage] at ($(nli.north)+(0,2.5mm)$) {VERIFICATION / GATING};
\node[stage, above=1.5mm of expand]  {EXPANSION};
\node[stage, above=1.5mm of enrich]  {ENRICH};
\node[stage, above=1.5mm of out]     {OUTPUT};

\draw[ar] (corpus) -- (extract);
\draw[ar] (extract.east) -- ++(3mm,0) |- (nli.west);
\draw[ar] (extract.east) -- ++(3mm,0) |- (sim.west);
\draw[ar] (nli.east) -- ++(3mm,0) |- (expand.west);
\draw[ar] (sim.east) -- ++(3mm,0) |- (expand.west);
\draw[ar] (expand) -- (enrich);
\draw[ar] (enrich) -- (out);
\draw[arg] (out.south) |- (val.east);
\end{tikzpicture}}
\caption{The \sysname{} construction pipeline. Extracted claim
sentences are routed down one of two pathways: claim-driven relations
are verified by NLI entailment against local in-paper context, while
similarity-based relations, which have no single seed sentence, are
gated by abstract cosine similarity and year-gap constraints. The
reliability audit (Section~\ref{sec:pilot}) evaluate the released graph }
\label{fig:graph_const}
\end{figure*}

We constructed \sysname{} using a five-stage pipeline shown in Figure~\ref{fig:graph_const}. For the four
claim-based relations (causal, limitation, future-direction, and dispute),
we use NLI~\cite{he2021debertav3} to verify whether the
claims are supported. The two similarity-based relations
(\emph{direct extension} and \emph{temporal semantic}) are identified
using abstract-level cosine similarity and publication year-gap
constraints.

\begin{table*}[t]
\centering
\small
\setlength{\tabcolsep}{4pt}
\renewcommand{\arraystretch}{1.15}
\caption{The \sysname{} relation schema with representative claim
sentences examples}
\begin{tabular}{@{}p{0.075\textwidth}p{0.115\textwidth}p{0.30\textwidth}p{0.40\textwidth}@{}}
\toprule
\textbf{Type} & \textbf{Sub-type} & \textbf{Description} & \textbf{Example claim sentence} \\
\midrule

\multirow{10}{*}{\parbox{0.075\textwidth}{\raggedright Claim-\\driven\\ \emph{(NLI-\\verified)}}}
& \emph{causal extension}
& A result or method established in one paper causally enables a
  development in the other.
& ``\textit{convolution kernels are not good at modeling long-range
  dependencies of image contents and features, because they only
  process a local neighborhood, either in space or time.}''
   \\[4pt]

& \emph{limit addressed}
& One paper addresses a limitation the other explicitly acknowledges
  about its own method or results.
& ``\textit{A limitation of our dataset is that we were limited to Pew
  research\ldots}''  \\[4pt]

& \emph{future realized}
& One paper realizes a future direction the other proposes but does
  not pursue.
& ``\textit{As discussed above, it would be an interesting direction
  to investigate a shift-variant CNN.}''
   \\[4pt]

& \emph{dispute}
& One paper disputes a claim or result of the other.
& ``\textit{We challenge many of the assumptions underlying this
  work\ldots}''  \\
\midrule

\multirow{5}{*}{\parbox{0.075\textwidth}{\raggedright Similarity-\\based}}
& \emph{direct extension}
& One paper extends a method or result of the other, with no stated
  limitation or future direction as the trigger.
& \emph{No single seed sentence; gated by abstract cosine
  $\geq 0.92$ and year-gap $\leq 2$ years.} \\[4pt]

& \emph{temporal semantic}
& One paper updates the semantic content of the other in a later time
  period.
& \emph{No single seed sentence; gated by abstract cosine
  $\geq 0.92$ and year-gap $\geq 5$ years.} \\
\bottomrule
\end{tabular}

\label{tab:relation_schema}
\end{table*}

\subsection{Relation Schema}
\label{sec:corpus_schema}
\sysname{} releases six active relation types. \textbf{direct extension}
captures cases where $s$ extends a method or result of $t$;
\textbf{future realized}, cases where $s$ realises a future direction
proposed by $t$; and \textbf{limit addressed}, cases where $s$
addresses a limitation noted in $t$. \textbf{causal extension} marks
edges where a result in $t$ causally enables a development in $s$;
\textbf{temporal semantic}, edges where $s$ updates the semantic
content of $t$ in a new time period; and \textbf{dispute}, edges where
$s$ disputes a claim or result of $t$.
The claim-extraction patterns and corresponding implementation details are provided with the released corpus. In Table~\ref{tab:relation_schema} , we provide a detailed description of each relation type and subtype, with examples of claim sentences.

\subsection{Corpus Construction}
\label{sec:corpus_pipeline}

\paragraph{Source data and venue filtering}
We start from the Semantic Scholar Open Research Corpus
\citep{lo2020s2orc} and filter to papers from three research
communities across 2015--2024:
\begin{itemize}[leftmargin=*,itemsep=2pt,topsep=2pt]
  \item \textbf{NLP} (13{,}461 papers): ACL, EMNLP, NAACL,
  COLING, EACL, TACL, AACL, LREC.
  \item \textbf{ML} (9{,}344 papers): NeurIPS.
  \item \textbf{Vision} (9{,}754 papers): CVPR, ICCV,
  WACV, ACCV.
\end{itemize}
Papers are filtered to abstract length $\geq 300$ characters, and
sampling targets $\sim 1500$ papers per (category, year) cell. The
final corpus contains 32{,}559 papers. We treat 2019--2024 as the
corpus's analysis window for cross-venue findings
(Section~\ref{sec:findings}).

\paragraph{Claim extraction}
For each paper, we segment the abstract and section text into sentences.
We then identify candidate claims using case-insensitive rule-based
lexical patterns for five categories: \emph{contributions},
\emph{limitations}, \emph{future directions}, \emph{disputes}, and
\emph{causal claims}. To limit over-representation from individual
papers, we apply category-specific per-paper caps and retain candidate
sentences between 20 and 500 characters.

\paragraph{NLI verification of claim-driven relations}
In this stage, we verify whether candidate claim-driven relations are supported by their surrounding textual context. We formulate this as a natural language inference task, using a DeBERTa-v3-MNLI \citep{he2021debertav3} model to assess whether each extracted claim is entailed by its local in-paper context.
Only claims receiving an entailment prediction are retained, reducing unreliable relations arising from negation, hedging, or rhetorical framing. While NLI has been widely applied to factual claim verification~\cite{wadden2020fact,thorne2018fever}, \sysname{} extends this principle to citation-based relational claims, using contextual support as an additional criterion for constructing reliable scholarly relations.

\paragraph{Relation expansion and graph enrichment}
The final stage transforms the validated relation evidence into a structured scholarly graph. Claim-driven relations are extended from NLI-verified seeds to semantically related paper pairs, while \emph{direct extension} and \emph{temporal semantic} are inferred directly from strong semantic similarity combined with distinct temporal constraints. 
Across all relation types, this process reduces a large candidate space to 573,126 typed edges, yielding a graph that captures multiple forms of continuity, response, and development across the literature.

\paragraph{Graph enrichment} The graph is subsequently enriched with thematic and contextual information. Papers are organized into latent topical groups derived from their abstract representations, while edges retain complementary semantic, temporal, topical, and confidence signals. These attributes support downstream analyses of both the structure and strength of relationships within the resulting scholarly network.

\subsection{Corpus Statistics}
\label{sec:corpus_stats}

\begin{table}[t]
\caption{\sysname{} corpus statistics: 32{,}559 papers across
NLP, ML, and Vision (2015--2024), 573{,}126 typed edges
across six relations, and 90{,}020 NLI-verified claim seeds.}
\label{tab:corpus_stats}
\centering
\small
\begin{tabular}{lr}
\toprule
\textbf{Papers} & 32{,}559 \\
\textbf{Edges (verified)} & 573{,}126 \\
\textbf{Year range} & 2015--2024 \\
\midrule
\multicolumn{2}{l}{\textbf{NLI-verified claim seeds}} \\
\quad causal              & 59{,}832 (86.4\%) \\
\quad limitation          & 23{,}159 (88.0\%) \\
\quad future-direction    &  3{,}669 (87.8\%) \\
\quad dispute             &  3{,}360 (90.9\%) \\
\quad \textbf{Total}      & \textbf{90{,}020} \\
\midrule
\multicolumn{2}{l}{\textbf{Edges per relation type}} \\
\quad \emph{causal extension}      & 202{,}908 (35.4\%) \\
\quad \emph{limit addressed}       & 190{,}243 (33.2\%) \\
\quad \emph{temporal semantic}     & 101{,}977 (17.8\%) \\
\quad \emph{future realized}       &  43{,}666 (7.6\%) \\
\quad \emph{direct extension}      &  26{,}001 (4.5\%) \\
\quad \emph{dispute}                &   8{,}331 (1.5\%) \\
\midrule
\multicolumn{2}{l}{\textbf{Per-category papers}} \\
\quad NLP (ACL, EMNLP, NAACL,...)     & 13{,}461 (41.3\%) \\
\quad ML (NeurIPS)                  &  9{,}344 (28.7\%) \\
\quad Vision (CVPR, ICCV, WACV, ACCV)  &  9{,}754 (30.0\%) \\
\midrule
\multicolumn{2}{l}{\textbf{Trajectories (strict, 4 progression rels)}} \\
\quad Total ($\geq 3$)              & 287{,}284{,}907 \\
\quad Coverage                      & 63.5\% \\
\midrule
\multicolumn{2}{l}{\textbf{Trajectories (inclusive, 6 rels)}} \\
\quad Total ($\geq 3$)              & 287{,}870{,}035 \\
\quad Coverage                      & 72.8\% \\
\bottomrule
\end{tabular}
\end{table}

Table~\ref{tab:corpus_stats} reports basic statistics. The temporal
split places early years in training (citing-year $\leq 2020$),
2021-2022 in validation, and 2023--2024 in test. Causal extension is the largest single relation (35.4\%), followed by limit addressed (33.2\%); together the four claim-driven relations account for 77.7\% of edges. The corpus is roughly balanced between NLP (41.3\%), Vision (30.0\%) and  ML (NeurIPS) slice (28.7\%). \sysname{} additionally admits 287M
length-$\geq 3$ progression trajectories covering 72.8\% of papers
(Table~\ref{tab:traj_both_defs}),
which motivates the chain
validation in section~\ref{sec:pilot}.

\begin{table}[h]
\caption{Trajectory counts under both definitions.}
\label{tab:traj_both_defs}
\centering
\small
\begin{tabular}{lrr}
\toprule
Length & Strict & Inclusive \\
\midrule
3      &   8{,}495{,}704 &   9{,}453{,}303 \\
4      & 100{,}928{,}858 & 101{,}628{,}225 \\
5      & 176{,}560{,}345 & 176{,}788{,}507 \\
\midrule
\textbf{Total ($\geq 3$)} & \textbf{285{,}984{,}907} & \textbf{287{,}870{,}035} \\

\bottomrule
\end{tabular}
\end{table}

\section{Human Validation}
\label{sec:pilot}
\label{app:pilot_details}

To assess the reliability of \sysname{}'s typed-edge labels beyond
the NLI verifier, we conducted a pilot study of a 3 human annotator, 2 PhD students and one master's student,  on a 
sample of 520 items: 300 edges (600 papers) for relation validation, 150 edges
for temporal validation, and 70 trajectories (50 length-3, 20
length-4) for chain validation. The 300 relation-validation edges
were sampled proportionally across the six relation types and
across three year buckets. The detailed result is shown in Table~\ref{app:pilot_details}.

\subsection{Annotation Protocol}

Each annotator received an Excel workbook with four sheets: a README,
three annotation sheets for relation, temporal, and chain judgments.
For each of the 300 sampled edges, annotators were shown the source
and target paper titles and years, the assigned relation type, and the
claim sentence linked to the edge. For causal edges, they also saw the
surrounding context and the NLI entailment probability. Annotators then
choose one of four labels:

\begin{itemize}[leftmargin=*,itemsep=2pt,topsep=2pt]
  \item \textsc{correct}: the relation type is supported by the evidence;
  \item \textsc{partially correct}: the relation is partly right, but
  another label could also fit;
  \item \textsc{incorrect}: another relation type, or no relation, fits better;
  \item \textsc{unclear}: the evidence is missing or ambiguous.
\end{itemize}

\subsection{Inter-Annotator Agreement}

\begin{table}[t]
\caption{Inter-annotator agreement across the three pilot tasks.
Fleiss' $\kappa$ values in $[0.61, 0.80]$ are conventionally
interpreted as ``substantial'' agreement.}
\label{tab:iaa_results}
\centering
\small
\begin{tabular}{lc}
\toprule
Metric & Value \\
\midrule
\multicolumn{2}{l}{\emph{Relation validation} ($n = 300$):} \\
$\;\;$Fleiss' $\kappa$              & 0.74 \\
$\;\;$Cohen's $\kappa$ (mean pairwise) & 0.75 \\
$\;\;$Unanimous / majority / 1-1-1     & 75 / 22 / 3\% \\
\midrule
\multicolumn{2}{l}{\emph{Temporal validation} ($n = 150$):} \\
$\;\;$Fleiss' $\kappa$ (gap plausibility)             & 0.68 \\
$\;\;$Fleiss' $\kappa$ (semantic-update correctness)  & 0.60 \\
\midrule
\multicolumn{2}{l}{\emph{Chain validation} ($n = 70$):} \\
$\;\;$Cohen's $\kappa$ (mean pairwise, binary coherent) & 0.73 \\
$\;\;$Mean overall coherence (1--5 Likert)              & 3.81 \\
\bottomrule
\end{tabular}
\end{table}

Relation-validation Fleiss' $\kappa = 0.74$ matches the SciFact
benchmark \citep{wadden2020fact} and exceeds FEVER
\cite{thorne2018fever}($\kappa = 0.68$) for comparable 
claim-verification tasks. Agreement is lowest on temporal-semantic
correctness ($\kappa = 0.61$), reflecting the inherent difficulty
of distinguishing methodologically similar papers across different
time periods.

\subsection{Pipeline Precision by Relation Type}

\begin{table}[t]
\caption{Full per-relation validation breakdown. ``\% valid''
counts \textsc{correct} judgments; ``\% partial'' counts
\textsc{partially correct} judgments; ``\% invalid'' counts
\textsc{incorrect} or \textsc{unclear}. All are by 3-annotator
majority.}
\label{tab:per_relation_full}
\label{tab:per_relation_precision}
\centering
\footnotesize
\setlength{\tabcolsep}{2.5pt}
\begin{tabular}{lcccccc}
\toprule
Relation & $n$ & $\kappa$ & valid\% & partial\% & invalid\% & unclear\% \\
\midrule
\emph{direct extension}   &  15 & 0.82 & 87\% &  7\% &  7\% &  0\% \\
\emph{limit addressed}    &  27 & 0.81 & 85\% &  7\% &  4\% &  4\% \\
\emph{causal extension}   & 185 & 0.74 & 82\% &  9\% &  6\% &  3\% \\
\emph{future realized}    &  34 & 0.71 & 79\% & 12\% &  6\% &  3\% \\
\emph{temporal semantic}  &  24 & 0.61 & 65\% & 21\% &  8\% &  6\% \\
\emph{dispute}             &  15 & 0.79 & 63\% & 17\% & 20\% &  0\% \\
\midrule
\textbf{Overall}           & 300 & 0.74 & 79.9\% & 11\% &  8\% &  3\% \\
\bottomrule
\end{tabular}
\end{table}

Three-annotator majority agreement matches the pipeline label on 79.9\% of relation-validation edges, ranging from 87\% for \emph{direct extension} to 63\% for \emph{dispute}. The four NLI-verified claim-driven relations (\emph{causal}, \emph{limitation}, \emph{future direction}, and \emph{dispute}) reach 77\% mean precision, while the two similarity-only relations (\emph{direct extension} and \emph{temporal semantic}) reach 76\% mean precision (Table~\ref{tab:per_relation_full}).

\emph{Temporal semantic} has the second-lowest precision (65\%);
annotators frequently reclassified these as \emph{direct extension},
suggesting the schema distinction itself is contested rather than the
annotation noisy. \emph{Dispute} shows the opposite pattern: high
agreement ($\kappa = 0.79$) but only 63\% \textsc{correct}, because
``in contrast to''/``unlike [prior work]'' phrasing is typically used
to describe methodological differences rather than genuine dispute
($\sim$37\% of dispute edges).

\subsection{Trajectory Coherence}

70\% of the 70 sampled trajectories are rated coherent (3-annotator
majority overall Likert $\geq 4$), with mean overall coherence
3.81/5. Length-3 chains are more coherent than length-4 (73\% vs
65\%) as the probability of one weak transition compounds over
longer paths. The cleanest patterns are
\emph{future realized $\to$ direct extension $\to$ limit addressed}
(88\% coherent) and \emph{limit addressed $\to$ direct extension
$\to$ future realized} (83\%), representing canonical
"future-work-realised-then-limited" and
"limitation-addressed-then-extended" research narratives.These coherence rates suggest that
\sysname{} captures meaningful research trajectories rather than random
edge combinations (Table~\ref{tab:per_pattern_chain}).

\begin{table}[h]
\caption{Coherence rates broken down by edge-type pattern for the
70 sampled chains. ``Coherent'' = majority overall Likert $\geq 4$.}
\label{tab:per_pattern_chain}
\centering
\small
\setlength{\tabcolsep}{3pt}
\begin{tabular}{lcc}
\toprule
Pattern & $n$ & coherent\% \\
\midrule
\emph{future real $\to$ direct ext $\to$ limit addr}         &  8 & 88\% \\
\emph{limit addr $\to$ direct ext $\to$ future real}         &  6 & 83\% \\
\emph{future real $\to$ causal ext $\to$ direct ext}         &  5 & 80\% \\
\emph{direct ext $\to$ limit addr $\to$ future real}         &  4 & 75\% \\
\emph{direct ext $\to$ causal ext $\to$ dispute}              &  7 & 71\% \\
\emph{causal ext $\to$ direct ext $\to$ limit addr}          &  9 & 67\% \\
Other length-3 patterns                                          & 11 & 64\% \\
Length-4 chains (all patterns)                                  & 20 & 65\% \\
\midrule
\textbf{Overall}                                                & 70 & 70\% \\
\bottomrule
\end{tabular}
\end{table}

\subsection{Temporal Validation Results}

We found 79\% of cases follow temporal plausibility. The 22 borderline cases are mostly future realized edges with
short year gaps of 1-2 years. Annotators noted that this fast
realisation can happen in rapidly moving NLP areas, but it is still
faster than what is usually expected for future work follow-up.
The other 9 cases were implausible, mostly being direct extension edges with intervals between years greater than 10 years. These were often considered temporal semantic edges or unrelated edges.

Among temporal semantic relationships, 64\% of citations are considered legitimate temporal updates. The other 22\% have been reallocated into direct extension category. These citations refer to the method presented by the preceding publication but do not update the method temporally.

\section{Methods}
\label{sec:methods}

We benchmark five models that form a controlled ablation chain over
four representational axes: \emph{graph structure}, \emph{pair-level
scoring}, \emph{engineered structural features}, and
\emph{classifier family}. Each successive model adds exactly one
factor beyond its predecessor, so the AUC delta at each step
isolates the contribution of that factor
(Section \ref{sec:methods_chain}). We also evaluate encoder-based embeddings using four cluster-validity
metrics. We introduce a new \emph{temporal-coherence} (TC) diagnostic metric, which tests whether publication years are captured in the representation space
(Section \ref{sec:methods_metrics}).

\subsection{Task Formulation}
\label{sec:methods_task}

\sysname{} supports two prediction tasks.

\paragraph{Typed link prediction}
For each candidate directed paper pair $(s,t)$, where $s$ is published
no earlier than $t$, we predict whether a typed citation edge exists.
This is treated as a binary classification task. Evaluation uses hard negative examples sampled from a topic-year-stratified pool. The pair representation combines SPECTER2 embeddings, neighbourhood statistics from the time-truncated graph, and temporal metadata.

\paragraph{Typed relation classification}
Given that an edge $(s, t)$ exists, predict its relation type
$r \in R = \{\emph{direct extension}, \emph{future realized},
\emph{limit addressed},\emph{dispute}, \\ \emph{causal extension},\emph{temporal semantic} \}$ as a six-way
classification problem. 

For both tasks, edges are split temporally: training edges have
citing year $\leq 2020$, validation $\in \{2021, 2022\}$, and test
$\in \{2023, 2024\}$. All structural features visible to the model
at inference time on a candidate edge with citing year $y$ are
computed using only edges with citing year $< y$, which prevents leakage of test-period neighbourhood statistics into the feature
representation (Section \ref{app:training_protocol}).

\subsection{Five-Model Ablation Chain}
\label{sec:methods_chain}

We evaluate five models that progressively introduce different sources of information and modeling capacity. \textsc{SPECTER2-kMeans} serves as a content-only baseline using pretrained paper representations and cosine similarity. \textsc{SPECTER2-Agg} extends this baseline by incorporating information from typed citation neighbourhoods. \textsc{Pair-MLP-Base} moves from representation-level similarity to learned pairwise prediction using paper-pair embeddings together with basic structural signals. \textsc{SciTraj-Pair}, our primary model, further incorporates the full set of structural, typed-relation, topical, and temporal features described in Section~\ref{sec:methods_features}. Finally, \textsc{SciTraj-GBM} uses the same feature set with a gradient-boosted tree model, allowing us to separate the contribution of the engineered features from that of the neural classifier. Together, this sequence provides a controlled comparison of content, graph context, pairwise modeling, engineered relational features, and classifier choice.

\subsection{Pair Features Construction}
\label{sec:methods_features}

The 48-dim feature vector consumed by \textsc{SciTraj-Pair} and
\textsc{SciTraj-GBM} groups into Per-relation neighborhood cosine features (6),
per-relation log-degrees of source and target (24),
common-neighbour features - Adamic--Adar
\citep{adamic2003friends}, Jaccard, preferential attachment, and
related counts (6), topic-match indicator (2), year-gap basis (5),
edge-structure flags (2), and source-target degree ratios (3). All
features are standardised to fit on the
training pair distribution. Clear definitions and the per-feature
ablation appear in Section~\ref{sec:ablation}.

\subsection{Training Protocol}
\label{sec:methods_training}
\label{app:training_protocol}

Pair-MLPs train for 20 epochs with Adam at $10^{-3}$ and dropout
$0.3$; \textsc{SciTraj-GBM} uses LightGBM defaults with early
stopping on validation AUC. \textsc{SciTraj-Pair} is reported as
multi-seed mean $\pm$ std across five seeds; other models are
single-seed. All experiments fit within a single A100 GPU.

\textsc{SPECTER2-Agg} replaces each paper's raw SPECTER2 embedding
$e_p$ with a neighbour-aggregated representation:
\begin{equation}
  h_p = \alpha\, e_p + (1 - \alpha) \cdot
  \frac{1}{\sum_{q \in \mathcal{N}(p)} w_{pq}}
  \sum_{q \in \mathcal{N}(p)} w_{pq}\, e_q,
\end{equation}
with $\alpha = 0.5$ and $w_{pq}$ the edge-weight attribute of
edge $(p, q)$. The neighbourhood $\mathcal{N}(p)$ pools across all
typed neighbours of $p$ (in-edges and out-edges treated
symmetrically for this baseline). Pair scoring on the aggregated
representations uses cosine similarity.

For a candidate edge with citing year $y$, we compute structural
features using only edges from earlier years, that is, edges with
citing year $< y$. For training edges, features are computed on the
training subgraph with citing year $\leq 2020$. For validation and
test edges, the visible graph includes all training edges and any
edges that occur before the candidate edge's citing year. Topic
features and SPECTER2 cosine scores use static node features, so they
are not changed by this time truncation.

For each positive edge, we sample one hard negative from the same k-means topic cluster as the source paper. The negative edge has a year gap within
$\pm 2$ years of the gold edge and is not connected to the source paper
$s$ in the visible subgraph. Negatives are resampled at each epoch. For
downstream Tasks A and B, we use a separate negative-sampling protocol
based on SPECTER2-nearest non-edges
(Section \ref{sec:downstream_a}-\ref{sec:downstream_b}).

\textsc{SciTraj-GBM} uses LightGBM \citep{ke2017lightgbm} with
default hyperparameters  and early stopping after 10 rounds without improvement in validation AUC

\textsc{SciTraj-Pair} is reported as mean $\pm$ standard deviation
across five seeds $\{42, 17, 2024, 511, 7\}$ for the headline AUC
in Table~\ref{tab:lp_results}. The ablation chain entries for
\textsc{SPECTER2-kMeans}, \textsc{SPECTER2-Agg}, \textsc{Pair-MLP-Base},
and \textsc{SciTraj-GBM} are reported single-seed (seed 42)
because their AUC variance under reshuffling is well below the
$+0.029$ headline delta. Feature-ablation runs
(Section \ref{sec:ablation}) are single-seed for the same reason.

\subsection{Cluster-Validity and Temporal-Coherence Metrics}
\label{sec:methods_metrics}

We define four metrics on the embedding spaces produced by
encoder-based models (pair-level scorers do not produce node
embeddings):
(i) \textbf{silhouette coefficient}
\citep{rousseeuw1987silhouettes} against $k$-means clusters
($k = 30$);
(ii) \textbf{Calinski--Harabasz index}
\citep{calinski1974dendrite} over the same partition;
(iii) \textbf{year-rank coherence} $\rho$: Spearman correlation
between embedding distance $d_{st} = 1 - \cos(h_s, h_t)$ and year
gap $|\Delta y_{st}|$ on test edges, in the spirit of
\citet{blei2006dynamic}; and
(iv) \textbf{temporal coherence (TC, ours)}: mean within-cluster
year variance normalised by global year variance,
\begin{equation}
  \mathrm{TC}(\mathbf{H}) = \frac{1}{|\mathcal{C}|}
     \sum_{c \in \mathcal{C}}
     \frac{\mathrm{Var}\bigl(\{y_p : p \in c\}\bigr)}
          {\mathrm{Var}\bigl(\{y_p : p \in \mathcal{P}\}\bigr)},
  \label{eq:tc}
\end{equation}
with $\mathcal{C}$ the $k=30$ $k$-means clusters and $y_p$ the
publication year of paper $p$. Lower is better: TC near $0$ means
each cluster is temporally homogeneous.

We treat TC as a \emph{falsifiability diagnostic metric}. Standard cluster-validity indices describe cluster
geometry, but they do not identify \emph{what semantic property} is
encoded by the clusters. Likewise, $\rho$ alone may conflate
cross-time citation behavior with publication-time information encoded
in the embeddings.

\paragraph{Falsifiability via year-shuffle}
Each metric is computed twice: once on the real graph $(a_r)$, and once
after permuting the publication-year array uniformly at random$(a_s)$. If
a metric truly captures temporal information, its value should
collapse under shuffling. \emph{A metric that does not drop under
shuffling is not measuring temporal information.}
Section~\ref{sec:year_shuffle} applies this protocol to the link
prediction models, where TC and $\rho$ collapse
across all encoders while silhouette and CH are unchanged.

\section{Results}
\label{sec:tier_comparison}

\subsection{Link Prediction}

\begin{table}[t]
\caption{Five-model link prediction comparison. $\Delta$ AUC is the
change from the predecessor row, isolating the contribution of the
single factor introduced.}
\label{tab:lp_results}
\centering
\small
\setlength{\tabcolsep}{4pt}
\begin{tabular}{lccc}
\toprule
Model & AUC & AP & $\Delta$ AUC \\
\midrule
\textsc{SPECTER2-kMeans}    & 0.876 & 0.835 & -- \\
\textsc{SPECTER2-Agg}       & 0.884 & 0.866 & $+0.008$ \\
\textsc{Pair-MLP-Base}      & 0.908 & 0.902 & $+0.024$ \\
\midrule
\textsc{SciTraj-Pair}       & \textbf{0.914 $\pm$ 0.005} & \textbf{0.895} & $+0.006$ \\
\textsc{SciTraj-GBM}        & 0.907 & 0.889 & $-0.007$ \\
\bottomrule
\end{tabular}
\end{table}

\begin{figure}[t]
\centering
\includegraphics[width=1\linewidth]{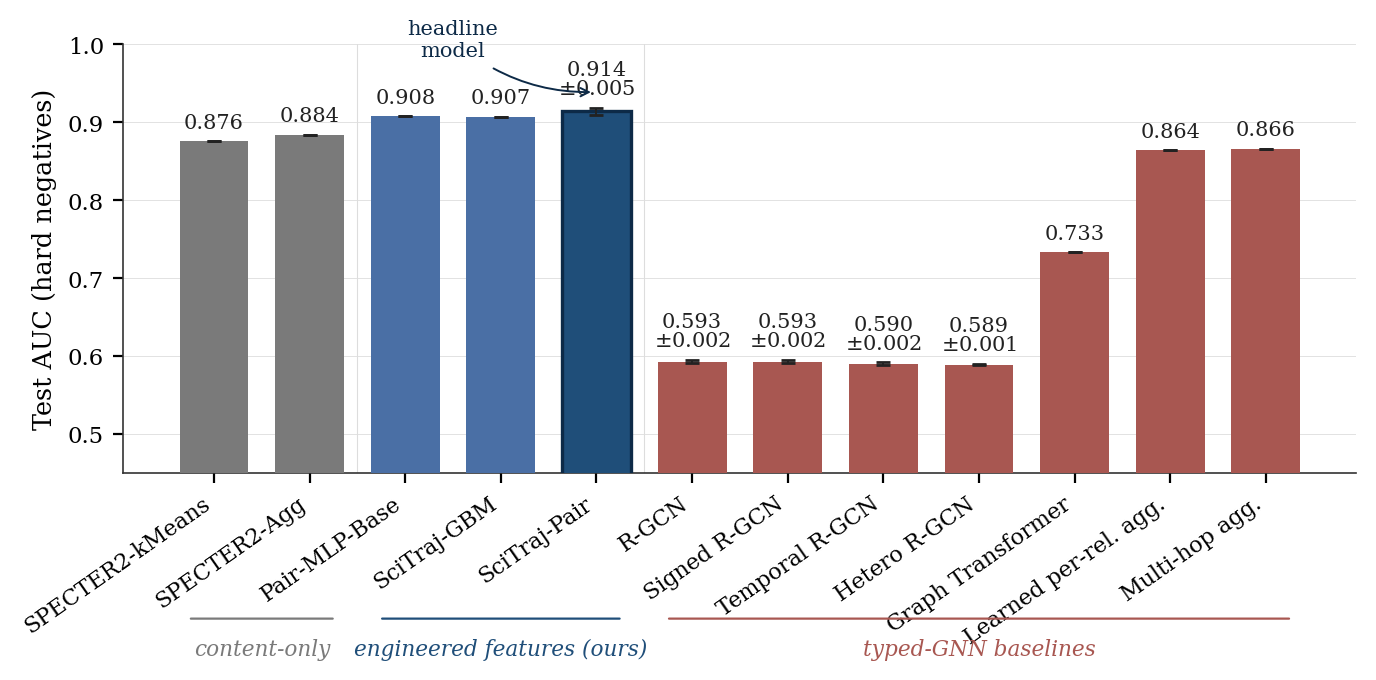}
\caption{Link-prediction AUC across three model families: content-only baselines (\textsc{SPECTER2-kMeans}, \textsc{SPECTER2-Agg}), engineered typed-graph features (\textsc{Pair-MLP-Base}, \textsc{SciTraj-Pair}, \textsc{SciTraj-GBM}), and typed-GNN baselines.}
\label{fig:tier_benchmark}
\end{figure}

The ablation chain decomposes the
gain step by step (in Figure~\ref{fig:tier_benchmark}): graph aggregation contributes $+0.008$ AUC,
pair-level scoring contributes $+0.024$ (the largest single jump),
and the engineered features contribute a further $+0.006$.
Swapping the Pair-MLP for LightGBM on identical features costs only
$-0.007$, which indicates the gain is in the input features rather
than the model family. Across five seeds, \textsc{SciTraj-Pair}
reaches AUC $0.9137 \pm 0.0051$, with a paired bootstrap test on
$10{,}000$ resamples against \textsc{SPECTER2-Agg} giving
$\Delta = +0.0294$, 95\% CI $[+0.0276, +0.0312]$, $p < 10^{-4}$.


We benchmarked seven typed-GNN architectures under identical hard
negatives: GraphSAGE with edge-type bias, R-GCN
\citep{schlichtkrull2018modeling}, temporal R-GCN, heterogeneous
multi-task R-GCN, Graph Transformer, learned per-relation
aggregation, and multi-hop aggregation. The four most directly
comparable (R-GCN, hetero GraphSAGE, temporal R-GCN, hetero
multi-task) collapse to AUC $\approx 0.59$; heavier architectures
recover to $0.73$--$0.87$ but remain below \textsc{Pair-MLP-Base}
($0.908$). \textsc{SciTraj-Pair} exceeds the best learned baseline
by $+0.048$ AUC and the canonical R-GCN by $+0.32$.


\subsection{Year-Shuffle Falsifiability Test}
\label{sec:year_shuffle}

A typed-graph model that scores well on temporal splits may still
be relying on time-correlated topical clusters rather than on
publication-time information itself. We test the distinction
directly with a year-shuffle protocol.

\begin{table}[t]
\caption{Year-shuffle falsifiability test.
\textsc{SciTraj-Pair} clears the $\epsilon = 0.05$ threshold.}
\label{tab:year_shuffle}
\centering
\small
\setlength{\tabcolsep}{3pt}
\begin{tabular}{lccc}
\toprule
Model & $a_r$ & $a_s$ & $\Delta$ \\
\midrule
\textsc{SPECTER2-kMeans} & 0.876 & 0.890 & $-0.014$ \\
\textsc{SPECTER2-Agg}    & 0.884 & 0.909 & $-0.025$ \\
\textsc{Pair-MLP-Base}   & 0.908 & 0.892 & $+0.016$ \\
\textsc{SciTraj-Pair}    & \textbf{0.914} & 0.626 & \textbf{+0.288} \\
\textsc{SciTraj-GBM}     & 0.907 & 0.892 & $+0.016$ \\
\bottomrule
\end{tabular}
\end{table}

\textsc{SciTraj-Pair} drops by $0.288$ AUC after shuffling, far
above $\epsilon$ (We pre-specify
$\epsilon = 0.05$ as the threshold: a model that depends on real publication times should show $\Delta = a_r - a_s \geq \epsilon$). All other models are stable or improve
slightly, indicating that their performance is driven by
year-invariant signals (text content, non-temporal graph
structure) rather than by publication-time information. The
same separation appears for the embedding-space diagnostics:
TC and $\rho$ collapse under shuffling, while silhouette and
Calinski--Harabasz are unchanged (Table~\ref{tab:year_shuffle}).

\subsection{Feature Ablation}
\label{sec:ablation}

Section ~\ref{sec:tier_comparison} established that \textsc{SciTraj-Pair}'s
48 engineered features are responsible for the headline gain over
content-based baselines. This section asks which of these features
carry the discriminative signal.

We aggregate the seven feature groups of
Section ~\ref{sec:methods_features} into four semantically meaningful
categories: \textbf{structural} (per-relation log-degrees and common-neighbour features, 36 dims, capturing network topology); \textbf{typed-relation} (per-relation neighbourhood cosine features, 6 dims, capturing relation-specific content alignment); \textbf{temporal} (year-gap basis, 5 dims, capturing publication chronology); and \textbf{topic} (topic-match indicator, 2 dims, capturing shared semantic neighbourhood).

\begin{table}[t]
\caption{Feature ablation on \textsc{SciTraj-Pair}. Each row removes
one feature group from the 48-feature input. ``Gain'' denotes the
category's contribution as a fraction of the chain gain.}
\label{tab:ablation}
\centering
\footnotesize
\setlength{\tabcolsep}{3.5pt}
\begin{tabular}{@{}lcc@{}}
\toprule
Removed group & $\Delta$ AUC & Gain \\
\midrule
Structural       & $-0.0412$ & 55\% \\
Typed relation   & $-0.0161$ & 22\% \\
Temporal         & $-0.0149$ & 22\% \\
Topic            & $-0.0046$ & 6\% \\
\midrule
Full \textsc{SciTraj-Pair} & 0.914 & -- \\
\bottomrule
\end{tabular}
\end{table}

Structural features contribute the most to \textsc{SciTraj-Pair}. Removing them lowers AUC by $0.0412$, which accounts for about 55\% of the gain over the content-only \textsc{SPECTER2-Agg} baseline. Typed-relation and year-gap features also improve performance, while topic-match features have only a small effect. Time is especially important. Removing temporal features causes a modest drop, but shuffling publication years reduces AUC by $0.288$. This shows that the model benefits from both temporal features and the overall chronological structure of the citation graph. Details report included in the Table~\ref{tab:ablation}.

\subsection{Downstream Tasks}
\label{sec:downstream}

We evaluate \sysname{} on three downstream tasks: \textbf{citation
augmentation},  \textbf{future citation prediction}, and \textbf{typed Relation Classification}. Across the three, we find a consistent pattern: typed-graph features dominate when the negative-sampling protocol is not tautologically aligned to either method.

\subsubsection{Citation Augmentation}
\label{sec:downstream_a}

For each held-out query paper $q$ (year $\geq 2022$, $\geq 3$ outgoing
edges) we rank candidate cited papers from a pool of gold citations
plus year-stratified hard negatives drawn as the top-200
SPECTER2-nearest non-cited papers within $\pm 2$ years of each gold
edge; evaluation on 500 queries.

\textsc{SciTraj-Pair} reaches R@10 of $0.367$ and MRR of $0.607$,
substantially below SPECTER2 cosine ($0.589$ and $0.876$).

SCITRAJ-PAIR and SPECTER2 are complementary scorers calibrated for different tasks: SPECTER2 wins on content-similarity-constrained retrieval, while typed-graph features dominate on the typed-prediction protocols of by margins of +0.6--0.7 AUC and +0.75 macro-F1.

\subsubsection{Future Citation Prediction}
\label{sec:downstream_b}

For each test year $Y \in \{2021, 2022, 2023, 2024\}$ we restrict
the training graph to citing-year $< Y$, mine hard negatives by
SPECTER2-nearest non-edges \citep{bhagavatula-etal-2018-content},
and retrain \textsc{SciTraj-Pair} per split.

\textsc{SciTraj-Pair} reaches AUC $0.89$--$0.94$ across years while
SPECTER2 cosine drops below random ($0.18$--$0.25$): the hard-negative
mining places cosine in an anti-correlated regime where positives
and negatives are competitively close in cosine space.

\subsubsection{Typed Relation Classification}
\label{sec:downstream_c}

Given a verified positive edge $(s,t)$, the task is to predict its
relation type on the 73{,}451-edge test split. We compare the
48-feature \textsc{SciTraj-Pair} representation with a 6-way softmax
head against three SPECTER2 baselines: a standard MLP, a class-weighted
MLP to handle the dominant \emph{causal extension} class, and
LightGBM.

\begin{table}[h]
\caption{Typed relation classification (six classes;
$n_{\text{test}} = 73{,}451$). Class-weighted and LightGBM
baselines added per reviewer feedback.}
\label{tab:downstream_c}
\centering
\small
\begin{tabular}{lc}
\toprule
Method & macro-F$_1$ \\
\midrule
Random (class prior)              & 0.164 \\
SPECTER2-MLP (unweighted)         & 0.193 \\
SPECTER2-MLP (class-weighted)     & 0.282 \\
SPECTER2-LightGBM                 & 0.260 \\
\textsc{SciTraj-Pair} features    & \textbf{0.948} \\
\bottomrule
\end{tabular}
\end{table}

\textsc{SciTraj-Pair} reaches macro-F$_1$ $0.948$, exceeding the
strongest SPECTER2 baseline by $+0.67$ (Table~\ref{tab:downstream_c}). The class-weighted
SPECTER2-MLP captures \emph{causal extension} ($F_1 = 0.73$) and
\emph{temporal semantic} ($F_1 = 0.58$) from text content but is
effectively blind to the other four relation types
($F_1 < 0.17$ for each). Text similarity captures topical overlap
but does not encode the citation-intent distinction between, say,
a future direction realised and a limitation addressed;
\textsc{SciTraj-Pair}'s typed neighbourhood structure does. The
remaining errors concentrate in \emph{direct extension}
$\leftrightarrow$ \emph{temporal semantic} and
\emph{causal extension} $\to$ \emph{dispute} pairs, both
semantically near-twin relations.

\section{Tracing Research Evolution}
\label{sec:findings}

A typed, claim-grounded citation graph supports questions that flat
citation lists cannot answer. We use \sysname{} to investigate two
questions about how scientific ideas move through the NLP, ML, and
Vision communities: (1) how cross-discipline citation patterns
compare to within-discipline patterns, and (2) which research topics
have emerged or declined across our 2015--2024 study window.

\subsection{Disciplinary Siloing in Typed Citation Flow}
\label{sec:findings_siloing}

Citation graphs naively suggest that adjacent disciplines exchange
ideas heavily. Our typed graph instead reveals
\emph{disciplinary siloing}: communities cite themselves more than
expected under uniform target choice, and cross-discipline citation
is below chance for every pair.

 For each (source category $c_s$, target category
$c_t$, year-bucket $b$), we compute two metrics over the 573{,}126
typed edges in \sysname{}: per-source rate (edges divided by source
paper count) and propensity (observed / expected under uniform
target choice). Propensity $>$~1 means over-representation relative
to a uniform-citation baseline; $<$~1 means under-representation.
Year buckets are 2015--2018 (NLP-and-ML era), 2019--2021 (Vision
joins), and 2022--2024 (LLM era).

\begin{table}[t]
\caption{Propensity (observed/expected) for typed citation flow,
2022--2024 period, on the similarity graph and the real S2 citation
graph. Diagonals are intra-venue.}
\label{tab:siloing_propensity}
\label{tab:real_citation_propensity}
\centering
\small
\setlength{\tabcolsep}{4pt}
\begin{tabular}{lcccccc}
\toprule
& \multicolumn{3}{c}{Similarity} & \multicolumn{3}{c}{Real S2} \\
\cmidrule(lr){2-4}\cmidrule(lr){5-7}
Src $\backslash$ Tgt & NLP & ML & Vis & NLP & ML & Vis \\
\midrule
NLP    & \textbf{2.07} & 0.41 & 0.10 & \textbf{1.99} & 0.56 & 0.06 \\
ML     & 0.19 & \textbf{2.71} & 0.48 & 0.38 & \textbf{2.26} & 0.65 \\
Vision & 0.07 & 0.86 & \textbf{2.41} & 0.11 & 1.12 & \textbf{2.12} \\
\bottomrule
\end{tabular}
\end{table}

Table~\ref{tab:siloing_propensity} shows a clear pattern of disciplinary siloing during 2022-2024, consistent across all three time buckets. Within-community citation is strongly over-represented, with NLP, ML, and Vision citing their own communities 2.1-2.7$\times$ more than expected under the uniform-target baseline. In contrast, nearly all cross-discipline flows fall well below baseline, with NLP$\leftrightarrow$Vision showing the strongest isolation. Vision$\to$ML is the only cross-community relation close to the expected level, reflecting that Vision is more closely connected to ML, while NLP remains comparatively isolated from the other two communities.

\subsection{Cross-Validation Against Real S2 Citations}
\label{app:real_citations}

The siloing finding in section \ref{sec:findings_siloing} is computed on
\sysname{}'s typed similarity graph. To cross-validate, we extract
the \emph{real citation subgraph}: for each of the 21{,}633 papers
in our corpus for which Semantic Scholar returns reference data, we filter their reference lists to keep only
references to other papers in our corpus, yielding 138{,}049
intra-corpus citation edges with median 4 edges per source. We
then compute propensity on this real-citation graph using identical
methodology (uniform target-choice baseline, same year buckets).

Three observations from Table~\ref{tab:real_citation_propensity}:
(i) all three diagonal propensities are above 1 in both methods
(real: 1.99-2.26; similarity: 2.07-2.71), that confirms the
intra-venue dominance pattern; (ii) NLP$\leftrightarrow$Vision is
the most isolated pair in both methods (real: 0.06/0.11;
similarity: 0.10/0.07); (iii) Vision$\to$ML is the strongest
cross-discipline link in both methods, and is in fact slightly
over-represented in the real-citation graph (propensity
1.12) where the similarity graph reports it at just below baseline
(0.86).
Real citations cross venue boundaries 35-100\% more often than our SPECTER2 similarity links. This is  likely because papers cite each other for reasons beyond methodological similarity. Even so, the overall directional pattern of siloing remains the same.

\subsection{Topic Emergence and the LLM Transition}
\label{sec:findings_emergence}

\sysname{}'s topic assignments allow us to track research topics
across 2019--2024 via an emergence ratio: 2022--2024 papers divided
by 2019--2021 papers per topic.
\begin{table}[t]
\caption{Top 5 emerging and declining topics by 2022--2024 / 2019--2021
paper-count ratio. Representative paper title is the nearest-centroid
paper of each topic.}
\label{tab:topic_emergence_top}
\centering
\small
\begin{tabular}{rlcl}
\toprule
$T$ & Cat & Ratio & Representative \\
\midrule
\multicolumn{4}{l}{\emph{Emerging}} \\
T29 & Vis & 2.70 & DiffMorpher: Diffusion model \\
T10 & NLP & 2.67 & wav2vec 2.0: Self-supervised speech \\
T6  & Vis & 2.53 & Learning Human Mesh Recovery \\
T20 & NLP & 2.39 & Distilling Step-by-Step (LLMs) \\
T14 & Vis & 2.26 & Spatio-Temporal Crop Aggregation \\
\midrule
\multicolumn{4}{l}{\emph{Declining}} \\
T26 & ML  & 0.51 & Greedy inference / lazy max-marginal \\
T12 & ML  & 0.58 & Empirical Risk Min.\ Under Fairness \\
T16 & ML  & 0.65 & MOReL: Model-Based Offline RL \\
T9  & ML  & 0.85 & Training Very Deep Networks \\
T7  & ML  & 0.88 & Neural Trees for Learning on Graphs \\
\bottomrule
\end{tabular}
\end{table}
Table~\ref{tab:topic_emergence_top} shows the top emerging and
declining topics. Vision dominates emerging: 7 of the top 10
fastest-growing topics are Vision (diffusion, 3D scenes,
self-supervised representation learning, video understanding,
multi-modal contrastive training). Three NLP topics also emerge,
including LLMs (T20, ratio 2.39$\times$), self-supervised speech (T10, 2.67$\times$), and parsing
(T3, 1.96$\times$). All top 5 declining topics are ML (classical
inference, fairness theory, offline RL, deep network training, graph
learning).

The siloing and emergence patterns are closely aligned. Vision is both the community most strongly connected to ML (Table~\ref{tab:siloing_propensity}) and the one associated with the fastest-growing topics (Table~\ref{tab:topic_emergence_top}). At the same time, classical ML topics are declining in absolute paper count, while Vision topics that build on ML methods such as diffusion models, self-supervised representation learning, and multimodal contrastive training are expanding rapidly.

\section{Conclusion}
\label{sec:conclusion}

We introduced \sysname{}, a typed, claim-grounded citation corpus of
32{,}559 papers and 573{,}126 directed edges across six relation
types. The four claim-driven relations are anchored to specific claim
sentences and verified by NLI entailment against in-paper context,
yielding 90{,}020 verified seeds across causal, limitation,
future-direction, and dispute relations. Using \sysname{}, we find two
broad patterns in typed research flow: research communities are
strongly siloed, with all 6 cross-community pairs falling below a
uniform target-choice baseline, and topic emergence from 2019--2024
concentrates in Vision and LLM-related work while several classical ML
topics decline. Both findings hold under a year-shuffle falsifiability
test, which also shows that publication time contributes signal beyond
year-correlated content in typed link prediction.
\sysname{} has several limitations that also point to directions for future work. The corpus focuses on selected NLP, ML, and Vision venues. So its coverage does not extend to the broader scientific literature but this framework can be adopted in other domains. In addition, some relation types are also difficult to separate consistently because they capture closely related forms of research development. Future work can broaden the corpus across disciplines and time periods, refine these relation boundaries, and there is a scope for strengthening contextual verification across the full relation schema.

We release the corpus, trained models, benchmark code, diagnostic
protocols, pilot annotations, and analysis code.
\sysname{}'s 287M typed length$\geq 3$ trajectories
 provide a foundation for forecasting
models that estimate plausible developments in a research line,
including next-edge prediction, trajectory completion, and
cross-community transfer. More broadly, \sysname{} shows that a typed,
claim-level view of citations can provide a more transparent basis for
studying how research evolves than homogeneous citation graphs alone.


\bibliographystyle{IEEEtranN}
\bibliography{custom}

\end{document}